\documentclass{article}
\usepackage{ijcai22}
\usepackage{multirow}
\usepackage{arydshln}
\usepackage{amssymb}

\usepackage{bm}
\usepackage{tikz}
\usepackage{times}
\usepackage{soul}
\usepackage{url}
\usepackage[hidelinks]{hyperref}
\usepackage[utf8]{inputenc}
\usepackage[small]{caption}
\usepackage{graphicx}
\usepackage{amsmath}
\usepackage{amsthm}
\usepackage{booktabs}
\usepackage{algorithm}
\usepackage{algorithmic}

\def\newcite#1{\citeauthor{#1}~\shortcite{#1}}

\title{Conversational Semantic Role Labeling with Predicate-Oriented Latent Graph}

\author {
    Hao Fei$^1$\and
    Shengqiong Wu$^1$\and
    Meishan Zhang$^2$\and
    Yafeng Ren$^{3*}$\And
    Donghong Ji$^1$\thanks{Corresponding author}
\affiliations
    $^1$Key Laboratory of Aerospace Information Security and Trusted Computing, Ministry  \\of Education, School of Cyber Science and Engineering, Wuhan University, China\\
    $^2$Institute of Computing and Intelligence, Harbin Institute of Technology (Shenzhen), China\\
    $^3$Laboratory of Language and Artificial Intelligence, Guangdong University of Foreign Studies, China
\emails
    \{hao.fei, whuwsq, renyafeng, dhji\}@whu.edu.cn, mason.zms@gmail.com
}

\begin{document}

\maketitle

\begin{abstract}
Conversational semantic role labeling (CSRL) is a newly proposed task that uncovers the shallow semantic structures in a dialogue text.
Unfortunately several important characteristics of the CSRL task have been overlooked by the existing works, such as the structural information integration, near-neighbor influence.
In this work, we investigate the integration of a latent graph for CSRL.
We propose to automatically induce a predicate-oriented latent graph (POLar) with a predicate-centered Gaussian mechanism, by which the nearer and informative words to the predicate will be allocated with more attention.
The POLar structure is then dynamically pruned and refined so as to best fit the task need.
We additionally introduce an effective dialogue-level pre-trained language model, CoDiaBERT, for better supporting multiple utterance sentences and handling the speaker coreference issue in CSRL.
Our system outperforms best-performing baselines on three benchmark CSRL datasets with big margins, especially achieving over 4\% F1 score improvements on the cross-utterance argument detection.
Further analyses are presented to better understand the effectiveness of our proposed methods.
\end{abstract}

\section{Introduction}

\begin{figure}[!t]
\centering
\includegraphics[width=1.0\columnwidth]{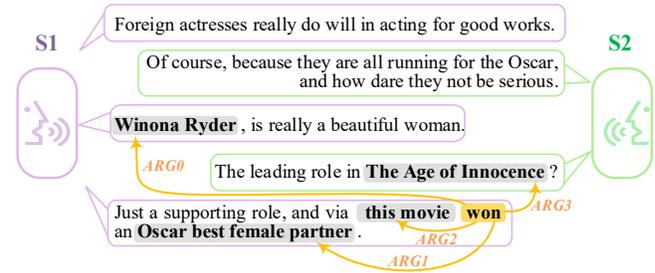}
\caption{
Illustration of conversational SRL by two speakers.
Word `won' in yellow background is the predicate, linking to its different types of arguments by arrows.
The arugments in the same utterance of the predicate are called intra-utterance arugment; those in different dialogue turns are marked as cross-utterance arugment.
}
\label{intro}
\end{figure}

Semantic Role Labeling (SRL) as a shallow semantic structure parsing task aims to find all the arguments for a given predicate
\cite{gildea-jurafsky-2000-automatic,marcheggiani-titov-2017-encoding,strubell-etal-2018-linguistically,fei-etal-2020-cross,FeiWRLJ21}.
Conversational SRL (CSRL) is a newly proposed task by \newcite{XuWSZSY21}, which extends the regular SRL into multi-turn dialogue scenario.
As illustrated in Fig. \ref{intro}, CSRL is characterized by that, the predicate is given at current utterance, while the correlated arguments are scattered in the history utterances of the dialogue that are generated by two speakers.
So far, few attempts have been made for CSRL \cite{XuWSZSY21,wu-etal-2021-csagn,wu-etal-2021-domain}, where, unfortunately, several key CSRL characteristics are still remained unexploted, which may hamper the further task improvements.

\textbf{First of all}, intuitively SRL structure echoes much with the syntactic dependency structure \cite{strubell-etal-2018-linguistically,marcheggiani-titov-2017-encoding}, and the existing regular SRL works frequently employ external structural information for performance enhancement, i.e., providing additional prior links between predicates and arguments.
However, it is quite intractable to directly employ the external syntax knowledge into CSRL for some reasons.
For examples, a dependency tree takes one single sentence piece as a unit, while a dialogue could contain multiple utterance sentences;
the parse trees from third-party parsers inevitably involve noises;
only a small part of the dependency structure can really offer helps, rather than the entire tree \cite{he-etal-2018-syntax}.
\textbf{Second}, the predicate-argument structures in CSRL are broken down and scattered into different utterances, which makes the detection of the CSRL more challenging.
Actually the chances are much higher for the predicate to find its arguments when they are being closer, i.e., near-neighbor influence.
In other words, nearer history utterances will show more impacts to the latest utterance.\footnote{Our data statistics shows that, cross-one-utterance arguments account for 60.3\% among all cross-turn arguments;
while the ratio decreases to 30.3\% for cross-two-utterance arguments.
}
Fig. \ref{intro} exemplifies the case.

\begin{figure}[!t]
\centering
\includegraphics[width=1.0\columnwidth]{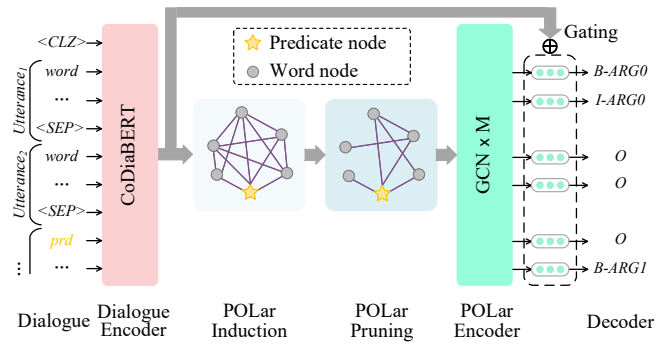}
\caption{
The overall CSRL framework.
}
\label{framework}
\end{figure}

Based on the above observations, in this paper we present an effective CSRL method with an innovative \underline{p}redicate-\underline{o}riented \underline{la}tent g\underline{r}aph (namely, POLar).
Unlike the explicit syntactic structures, we make use of a two-parameter \emph{HardKuma} distribution \cite{bastings-etal-2019-interpretable} to automatically induce latent graph from task's need (cf. $\S$\ref{Predicate-Oriented Latent Graph Induction}).
Particularly, we propose a predicate-centered Gaussian inducer for yielding the latent edges, by which the nearer and informative words to the predicate will be placed with more considerations.
The POLar is then dynamically pruned, so that only the task-relevant structure will be built, while the irrelevant edges are droped.
The overall CSRL framework is differentiable and performs predictions end-to-end (cf. Fig. \ref{framework}).

The BERT \cite{devlin2018bert} pre-trained language model (PLM) is extensively employed in existing works for CSRL performance boosts \cite{XuWSZSY21,wu-etal-2021-domain}.
Nevertheless, it could be problematic to directly leverage BERT for CSRL.
On the one hand, one entire dialog often consists of far more than two utterance sentences, while the raw BERT restricts the input with at maximum two sentence pieces, which consequently limits the PLM's utility.
Therefore, we consider adopting the DiaBERT \cite{liu-lapata-2019-text,li-etal-2020-modeling}, which is designed for well supporting multiple utterance inputs and thus yields better dialogue-level representations.
On the other hand, we note that in CSRL both two speakers use the personal pronoun in their own perspective (i.e., `I', `you'), and directly concatenating the multi-turn utterances into PLM will unfortunately hurt the speaker-role consistency, i.e., speaker coreference issue.
Therefore, we introduce a coreference-\underline{co}nsistency-enhanced DiaBERT (namely CoDiaBERT, cf. Fig. \ref{CoDiaBERT-fig}) that enhances the speaker-role sensitivity of PLM with a pronoun-based speaker prediction (PSP) strategy.

Our system significantly outperforms strong-performing baselines with big margins on three CSRL benchmarks.
In particular, over 4\% F1 score of improvement is achieved for detecting the cross-utterance type of arguments.
Further analyses reveal the usefulness of the proposed latent graph and the dynamic pruning method, as well as the CoDiaBERT PLM.
Also we show that our model effectively solves long-range dependence issue.
Overall, we make these contributions:

$\bullet$ We for the first time propose to improve the CSRL task by incorporating a novel latent graph structure.

$\bullet$ We construct a predicate-oriented latent graph via a predicate-centered Gaussian inducer.
The structure is dynamically pruned and refined for best meeting the task need.

$\bullet$ We introduce a PLM for yielding better dialogue-level text representations, which supports multiple utterance sentences, and is sensitive to the speaker roles.

$\bullet$ Our framework achieves new state-of-the-art CSRL results on three benchmark data.

\section{Related Work}

The SRL task aims at uncovering the shallow semantic structure of text, i.e. `\emph{who did what to whom where and when}'.
As a fundamental natural language processing (NLP) task, SRL can facilitate a broad range of downstream applications  \cite{shen-lapata-2007-using,liu-gildea-2010-semantic,wang-etal-2015-machine}.
By installing the current neural models, the current standard SRL has secured strong task performances \cite{strubell-etal-2018-linguistically,LiHZZZZZ19,0001ZLJ21}.
Recently, \newcite{XuWSZSY21} pioneer the task of CSRL by extending the regular SRL into multi-turn dialogue scenario, in which they provide benchmark datasets and CSRL neural model.
Later a limited number of subsequent works have explored this task \cite{wu-etal-2021-csagn,wu-etal-2021-domain}, where unfortunately several important features of CSRL are not well considered.
In this work, we improve the CSRL by fully uncovering the task characteristics.

This work also closely relate to the line of syntax-driven SRL \cite{marcheggiani-titov-2017-encoding,FeiRJ20,0001RJ20a}.
For the regular SRL, the external syntactic dependency structure is a highly-frequently equipped feature for performance enhancement, as the SRL shares much underlying structure with syntax \cite{he-etal-2018-syntax,0001RJ20,0001LLJ21}.
However, it could be problematic for CSRL to directly benefit from such convient syntactic knowledge, due to the dialogue nature of the text as we revealed earlier.
We thus propose to construct a latent structure at dialogue level, so as to facilitate the CSRL task with structural knowledge.
In recent years, constructing latent graph for downstream NLP tasks has received certain research attention \cite{ChoiYL18}.
As an alternative to the pre-defined syntactic dependency structure yielded from third-party parsers, latent structure induced from the task context could effectively reduce noises \cite{corro-titov-2019-learning}, and meanwhile enhance the efficacy (i.e., creating task-relevant connections) \cite{chen-etal-2020-inducing}.
In this work, we revisit the characteristic of CSRL, and based on the two-parameter Hard-Kuma distribution \cite{bastings-etal-2019-interpretable} investigate a predicate-oriented latent graph by proposing a predicate-centered Gaussian inducer.

\section{CSRL Framework}

\paragraph{Task modeling.}

Consider a conversation text $U$=$\{u_t\}_{t=1}^{T}$ ($T$ is the total utterance number), with each utterance $u_t$=$\{w_{0},w_{1},\cdots \}$ a sequence of words ($w_{0}$ is the utterance speaker).
In CSRL the predicate $prd$ is labeled as input at the current (lastest) utterance $u_T$.
We follow \newcite{XuWSZSY21}, modeling the task as a sequence labeling problem with a \emph{BIO} tagset.
CSRL system identifies and classifies the arguments of a predicate into semantic roles, such as \emph{A0}, \emph{A1}, \emph{AM-LOC}, etc, where we denote the complete role set as $\mathcal{R}$.
Given $U$ and the predicate $prd$, the system finally assigns each word $w$ a label $\hat{y}\in\mathcal{Y}$, where $\mathcal{Y}$=$(\{B,I\}$$\times$$\mathcal{R})\cup \{O\}$.

\paragraph{Framework overview.}

Our overall CSRL framework is illustrated in Fig. \ref{framework}.
The dialogue encoder first yields contextual representations for the input dialogue texts.
Then, the system generates the predicate-oriented latent graph (i.e., POLar induction), and performs structure pruning.
Afterwards, GCN layers encode the POLar into feature representations, based on which the predictions are finally made.

\begin{figure}[!t]
\centering
\includegraphics[width=1.0\columnwidth]{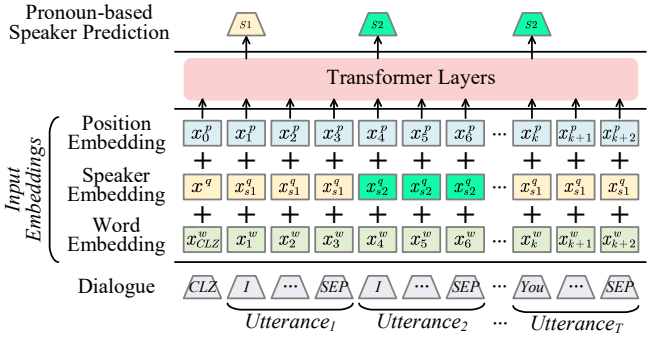}
\caption{
Illustration of the CoDiaBERT.
}
\label{CoDiaBERT-fig}
\end{figure}

\subsection{CoDiaBERT: Dialogue Encoder}

Contextualized word representations from BERT have brought great benefits to CSRL \cite{XuWSZSY21,wu-etal-2021-csagn,wu-etal-2021-domain}.
In this work, we follow them by borrowing the advances from PLM as well.
However, we notice that the raw BERT limits the input with maximum two sentence pieces, while often a conversation text can comprise far more than two utterance sentences.
Directly using BERT can thus lead to discourse information incoherency.
We thus leverage a dialogue-level BERT-like PLM \emph{DiaBERT} \cite{liu-lapata-2019-text}.
Technically, we pack the utterance with its speaker as a group, and concatenate those groups into a whole (separated with \emph{SEP} tokens), and feed into the PLM encoder.

The speaker coreference issue in conversational context may quite confuse the model.
For example, speaker \#1 would call speaker \#2 `you' in speaker \#1's utterance, while both speaker \#1 and speaker \#2 call themselves with the first-person pronoun `I'.
To strengthen the sensitivity of the speaker role, we further retrofit the DiaBERT so as to enhance the coreference consistency, i.e., CoDiaBERT.
Specifically, we based on the well-trained DiaBERT perform a \emph{pronoun-based speaker prediction} (PSP) upon DiaBERT, as shown in Fig. \ref{CoDiaBERT-fig}.
We first concatenate different utterance texts into a whole piece that are separated with $<$\emph{SEP}$>$ token.
Then we prepare three types of embeddings for each input token: 
1) word embedding $\bm{x}^w$, 
2) speaker id embedding $\bm{x}^q$,
and 3) position embedding $\bm{x}^p$,
all of which are fed into PLM for PSP:
\begin{equation}
\setlength\abovedisplayskip{2pt}
\setlength\belowdisplayskip{2pt}
\begin{aligned} \label{CoDiaBERT-prompt}
\bm{x}_i &= [\bm{x}^p ; \bm{x}^q ; \bm{x}^w]_i  \,, \\
\{ \cdots, \bm{h}_{i}, \cdots \}  &= \text{CoDiaBERT}^{\text{PSP}}( \{ \cdots, \bm{x}_i, \cdots \} ) \,.
\end{aligned}
\end{equation}
Based on the pronoun representation (i.e., the corresponding word is a pronoun), we encourage the PLM to predict the speaker id.

After PSP, the CoDiaBERT could yields better dialogue representations.
In our CSRL framework, CoDiaBERT will take as input the conversation texts (including the speaker id) as well as the predicate word annotation:
\begin{equation}
\setlength\abovedisplayskip{2pt}
\setlength\belowdisplayskip{2pt}
\begin{aligned} \label{CoDiaBERT-encoding}
\bm{x}_i &= [\bm{x}^p ; \bm{x}^q ; \bm{x}^w ; \bm{x}^{prd}]_i  \,, \\
\{ \cdots, \bm{h}_{i}, \cdots \}  &= \text{CoDiaBERT}^{\text{enc}}( \{ \cdots, \bm{x}_i, \cdots \} ) \,.
\end{aligned}
\end{equation}
where $\bm{x}^{prd}$ is the predicate binary embeddings $\{0,1\}$ indicating the presence or absence of the predicate word $prd$.
$\bm{h}_{i}$ denotes the output representation for the input token $w_i$.

\subsection{Latent Graph Encoder}

Based on the CoDiaBERT representation\footnote{
We abandon the special sentinel tokens (e.g., $<$\emph{CLZ}$>$ and $<$\emph{SEP}$>$), and only make use of the word and speaker tokens.} 
we can construct the POLar structure, which we will elaborate in the next section (cf. $\S$\ref{Predicate-Oriented Latent Graph Induction}).
In the POLar $\mathcal{G}=(V,E)$, each edge $\pi_{i,j} \in E$ is a real value that denotes a latent connection between node $v_i \in V$ to node $v_j \in V$ with a connecting intensity, i.e., $E$ is a $K \times K$ adjacent matrix ($|V|=K$).\footnote{The node could either be a word, or a speaker.}
Once we obtain the POLar we encode it into feature representations.
Specifically, we employ a multi-layer ($M$) graph convolutional network (GCN) \cite{marcheggiani-titov-2017-encoding}.
We denote the $m$-th layer of GCN hidden representation of node $v_i$ as $\bm{r}^m_i$:
\begin{equation}
\label{GCN}
\setlength\abovedisplayskip{2pt}
\setlength\belowdisplayskip{2pt}
\bm{r}_i^m = \text{ReLU}(\begin{matrix} \sum_{j=1}^K \end{matrix} \bar{A}_{i,j} \bm{W}^m_{1} \bm{r}_{j}^{m-1} / d_i + b^m ) \,,
\end{equation}
where $\bar{A}=E + I$ ($I$ is a $K \times K$ identity matrix),
$d_i=\sum_{j=1}^K E_{i,j}$ is for node normalization.
Note that the input of the initial layer is the CoDiaBERT representations, i.e., $\bm{r}_{i}^0=\bm{h}_{i}$
After total $M$ layers of message propagations, we expect the GCN can sufficiently capture the structural features.

\subsection{Decoder and Training}

To take the full advantages of the global dialogue contextual features, we create a residual connection from CoDiaBERT to the end of the GCN layer:
\begin{equation}
\setlength\abovedisplayskip{2pt}
\setlength\belowdisplayskip{2pt}
\bm{e}_{i} =  g_i \odot \bm{r}_i^M +   (1-g_i) \odot  \bm{h}_{i}  \,,
\end{equation}
where $\bm{e}_{i}$ is the final feature representation, which fuses both the contextual features and the structure-aware features.
$g_i$ is a gate mechanism that is learned dynamically:
\begin{equation} \label{gating}
\setlength\abovedisplayskip{2pt}
\setlength\belowdisplayskip{2pt}
g_i =  \sigma ( \bm{W}_2 \cdot [ \bm{r}_i^M ; \bm{h}_{i} ] ) \,.
\end{equation}
Based on $\bm{e}_{i}$ we adopt a Softmax classifier to
predict the labels for tokens:
\begin{equation}
\setlength\abovedisplayskip{2pt}
\setlength\belowdisplayskip{2pt}
\hat{y}_i = \text{Softmax} (\bm{e}_{i}) \,.
\end{equation}
Also the Viterbi algorithm is used to search for the highest-scoring tag sequence $\hat{\bm{Y}}=\{\hat{y}_1,\cdots,\hat{y}_K\}$.

Our training objective is to minimize the cross-entropy loss between the predictions $\hat{\bm{Y}}$ and the gold labels $\bm{Y}$.
\begin{equation}
\setlength\abovedisplayskip{3pt}
\setlength\belowdisplayskip{3pt}
\mathcal{L} = -\frac{1}{K} \begin{matrix} \sum_{j=1}^{K} \end{matrix} {y}_j \log \hat{y}_j \,,
\end{equation}
where $K$ is the total sequence length (i.e., $|V|$).

\section{Predicate-Oriented Latent Graph Induction}
\label{Predicate-Oriented Latent Graph Induction}

Since the goal of CSRL is to find the arguments of the predicate, it is crucial to treat the predicate word as the pivot and induce a predicate-oriented latent graph (POLar) to fully consider the near-neighbor influence.
Here we demonstrate how to develop the POLar structure.
First, we give a description on the theoretical fundamentation of the \emph{HardKuma} distribution, upon which we build the latent strucutre.
Then we introduce the predicate-centered Gaussian inducer.
Finally we present the method for dynamically pruning the POLar.

\begin{figure}[!t]
\centering
\includegraphics[width=1.0\columnwidth]{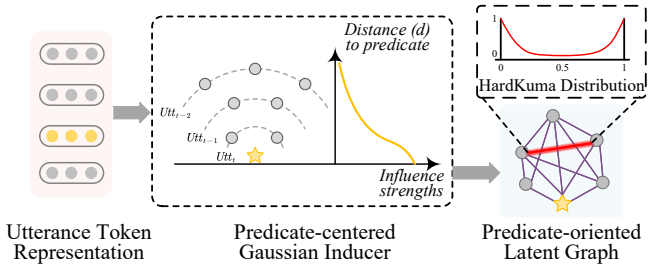}
\caption{
Induction of the predicate-oriented latent graph.}
\label{POLar-induction}
\end{figure}

\subsection{HardKuma Distribution}

HardKuma distribution \cite{bastings-etal-2019-interpretable} is derived from the \emph{Kumaraswamy} distribution (namely Kuma) \cite{1980A}, which is a two-parameters distribution over an open interval (0, 1), i.e., $t \sim \text{HardKuma} (a,b)$ where $a \in \mathbb{R}_{>0}$ and $b \in \mathbb{R}_{0}$ are the parameters controlling the shapes.
However, the Kuma distribution does not cover the two discrete points 0 and 1.
Thus, the HardKuma distribution adopts a \emph{stretch-and-rectify} method to support the closed interval of [0, 1].
This feature allows to predict soft connections probabilities between input words, i.e., a latent graph, where the entire process is fully differentiable.

First, we sample a variable from a ($0$,$1$) distribution, i.e., $U \sim \mathcal{U}(0,1)$, based on which we generate another variable from HardKuma's inverse CDF function:
\begin{equation}\label{HardKuma-1} 
\setlength\abovedisplayskip{2pt}
\setlength\belowdisplayskip{2pt}
k = \text{F}^{-1}_{K} (u,a,b) \,.
\end{equation}
Then we stretch the $k$ into $t$:
\begin{equation}\label{HardKuma-2} 
\setlength\abovedisplayskip{2pt}
\setlength\belowdisplayskip{2pt}
t = l + (r-l) *k \,,
\end{equation}
where $l <0$ and $r>1$ represent an open interval ($l$,$r$).\footnote{
Following the standard setup in \newcite{bastings-etal-2019-interpretable}, $l$=-0.1 and $r$=-1.1.
}
A Hard-Sigmoid function rectifies the $t$ into $h$ via
\begin{equation}\label{HardKuma-3} 
\setlength\abovedisplayskip{2pt}
\setlength\belowdisplayskip{2pt}
\text{F}^{-1}_{T} (t;a,b,l,r) = \text{F}_{K} (\frac{t-l}{r-l};a,b) \,.
\end{equation}
In short, we can summarize the HardKuma distribution as:
\begin{equation}\label{HardKuma-all} 
\setlength\abovedisplayskip{2pt}
\setlength\belowdisplayskip{2pt}
t \sim \text{HardKuma} (a,b,l,r) \,.
\end{equation}
For more technical details we refer the readers to the raw papers \cite{bastings-etal-2019-interpretable}.

\subsection{Predicate-centered Gaussian Inducer}

By sampling variables from \emph{HardKuma} distribution with trained parameters $a$ and $b$, we can generate the latent graph based upon the dialogue.
Specifically, we present a predicate-centered Gaussian inducer (PGI), so that the near neighbors to predicate that carry more important information would serve more contributions.

As depicted in Fig. \ref{POLar-induction}, we first upgrade each token representation into $\bm{h}^{'}_i$ with the prior of predicate word, via a predicate-centered Gaussian operator:
\begin{equation}
\setlength\abovedisplayskip{2pt}
\setlength\belowdisplayskip{2pt}
\begin{aligned} \label{PGI-1}
\bm{h}^{'}_i &= \text{PGI}(\bm{h}_i | \bm{h}_{i(prd)}) \,, \\
&= \frac{
f(d_{i,i(prd)}) \text{Softmax}(\frac{\bm{h}_i \cdot \bm{h}_{i(prd)}}{\sqrt{d_{i,i(prd)}}})
}{
\sum_l f(d_{i,l}) \text{Softmax}(\frac{\bm{h}_i \cdot \bm{h}_{l}}{\sqrt{d_{i,i(prd)}}})
} \,, 
\end{aligned}
\end{equation}
where $d=|i-i(prd)|$ is the edit distance between a token $w_i$ and the predicate $prd$.
Here $f(d)$ is a Gaussian distance, i.e., $f(d)=\exp(-\pi d^2)$.
So $\bm{h}^{'}_i$ is reduced into:
\begin{equation} \label{PGI-2}
\setlength\abovedisplayskip{2pt}
\setlength\belowdisplayskip{2pt}
\bm{h}^{'}_i = \text{Softmax}( -\pi d^2_{i,i(prd)} + \frac{\bm{h}_i \cdot \bm{h}_{l}}{\sqrt{d_{i,i(prd)}}} ) \,.
\end{equation}

Based on $\bm{h}^{'}_i$, we then create the parameter context representations (i.e., denoted as $\bm{s}^a$ and $\bm{s}^b$) via separate feedforward layers (i.e., $\bm{s}^{a/b}_i$=FNN$^{a/b}$($\bm{h}^{'}_i$)).
Then we build the prior parameter representations of the distribution:
\begin{equation}
\setlength\abovedisplayskip{2pt}
\setlength\belowdisplayskip{2pt}
\begin{aligned} \label{PGI-param}
\bm{a} &= \text{Norm}( \bm{s}^a_i (\bm{s}^a_j)^T  )  \,, \\
\bm{b} &= \text{Norm}( \bm{s}^b_i (\bm{s}^b_j)^T  )  \,.
\end{aligned}
\end{equation}
Thereafter, we can sample a soft adjacency matrix between tokens, i.e., $\pi_{i,j} \in E$:
\begin{equation} \label{sampler}
\setlength\abovedisplayskip{2pt}
\setlength\belowdisplayskip{2pt}
\pi_{i,j} = \text{HardKuma}( a_{i,j} , b_{i,j} , l ,r ) \,.
\end{equation}

\subsection{Dynamic Structural Pruning}

There are high chances that the induced POLar structure is dense, which would introduce unnecessary paths that are less-informative to the task need, i.e., noises.
Therefore, we adopt the $\alpha$-Entrmax \cite{correia-etal-2019-adaptively} to prune the POLar.
$\alpha$-Entrmax imposes sparsity constraints on the adjacency matrix $E$, and the pruning process automatically removes irrelevant information according to the contexts dynamically:
\begin{equation} \label{pruning}
\setlength\abovedisplayskip{3pt}
\setlength\belowdisplayskip{3pt}
E = \alpha\text{-Entrmax}( E ) \,,
\end{equation}
where $\alpha$ is a dynamic parameter controlling the sparsity.
When $\alpha$=2 the Entrmax becomes a Sparsemax mapping, while $\alpha$=1 it degenerates into a Softmax mapping.

\section{Experimentation}

\subsection{Setups}

We conduct experiments on three CSRL datasets \cite{XuWSZSY21}, including DuConv, NewsDialog and PersonalDialog, with average 10.1, 5.2 and 6.1 utterances per dialogue, respectively.
All the three data is in Chinese language.
We take the default data split as in \newcite{XuWSZSY21}, where DuConv has the 80\%/10\%/10\% ratio of train/dev/test, while NewsDialog and PersonalDialog are taken as out-of-domain test set.
Our CoDiaBERT shares the same architecture with the official BERT/DiaBERT (Base version), and is further post-trained on the CSRL data with PSP strategy.
GCN hidden size is set as 350.
We adopt Adam as the optimizer with an initial learning rate of 5e-4 with weight decay of 1e-5. 
The initial $\alpha$ value is 1.5.
To alleviate overfitting, we use a dropout rate of 0.5 on the input layer and the output layer. 

\begin{table*}[!t]
\centering
\resizebox{1.0\textwidth}{!}{
\begin{tabular}{llllllllll}
\toprule
\multirow{2}{*}{}&  \multicolumn{3}{c}{DuConv} & \multicolumn{3}{c}{NewsDialog}  & \multicolumn{3}{c}{PersonalDialog} \\
\cmidrule(r){2-4}\cmidrule(r){5-7}\cmidrule(r){8-10}
 & F1$_{all}$ & 	F1$_{cross}$ & 	F1$_{intra}$ & F1$_{all}$ & 	F1$_{cross}$ & 	F1$_{intra}$& F1$_{all}$ & 	F1$_{cross}$ & 	F1$_{intra}$ \\
\midrule
\multicolumn{10}{l}{$\bullet$  \bf w/ BERT}\\
\quad SimplePLM  \cite{SimpleBERT}$^*$ & 	86.54 & 	81.62 & 	87.02 & 	77.68 & 	51.47 & 	80.99 & 	66.53 & 	30.48 & 	70.00 \\
\quad CSRL \cite{XuWSZSY21}$^*$ & 	88.46 & 	81.94 & 	89.46 & 	78.77 & 	51.01 & 	82.48 & 	68.46 & 	32.56 & 	72.02 \\
\quad DAP \cite{wu-etal-2021-domain}$^\dag$ & 	89.97 & 	\underline{86.68} & 	90.31 & 	81.90 & 	\underline{56.56} & 	84.56 & \multicolumn{1}{c}{-} & \multicolumn{1}{c}{-} & \multicolumn{1}{c}{-} \\
\quad CSAGN \cite{wu-etal-2021-csagn}$^*$ & 	89.47 & 	84.57 & 	90.15 & 	80.86 & 	55.54 & 	84.24 & 	\underline{71.82} & 	\underline{36.89} & 	75.46 \\
\quad UE2E \cite{LiHZZZZZ19} & 	87.46 & 	81.45 & 	89.75 & 	78.35 & 	51.65 & 	82.37 & 	67.18 & 	30.95 & 	72.15 \\
\quad LISA \cite{strubell-etal-2018-linguistically} & 	89.57 & 	83.48 & 	91.02 & 	80.43 & 	53.81 & 	85.04 & 	70.27 & 	32.48 & 	75.70 \\
\quad SynGCN \cite{marcheggiani-titov-2017-encoding} & 	\underline{90.12} & 	84.06 & 	\underline{91.53} & 	\underline{82.04} & 	54.12 & 	\underline{85.35} & 	70.65 & 	34.85 & 	\underline{76.96} \\
\quad \bf POLar & 		\bf 92.06 & 		\bf 90.75 & 		\bf 92.64 & 		\bf 83.45 & 		\bf 60.68 & 		\bf 87.96 & 		\bf 73.46 & 		\bf 40.97 & 		\bf 78.02 \\
\hdashline
\multicolumn{10}{l}{$\bullet$ \bf w/ CoDiaBERT}\\
\quad SimplePLM \cite{SimpleBERT} & 	88.40 & 	82.96 & 	88.25 & 	79.42 & 	53.46 & 	82.77 & 	68.86 & 	33.75 & 	72.23 \\
\quad SynGCN \cite{marcheggiani-titov-2017-encoding} & 	\underline{91.34} & 	\underline{86.72} & 	\underline{91.86} & 	\underline{82.86} & 	\underline{56.75} & 	\underline{85.98} & 	\underline{72.06} & 	\underline{37.76} & 	\underline{77.41} \\
\quad \bf POLar & 		\bf 93.72 & 	\bf 	92.86 & 		\bf 93.92 & 		\bf 85.10 & 		\bf 63.85 & 		\bf 88.23 & 		\bf 76.61 & 		\bf 45.47 & 		\bf 78.55 \\
\bottomrule
\end{tabular}
}
\caption{Main results on three datasets.
Values with $*$ are copied from Wu \emph{et al}. [2021b];
with $\dag$ are copied from Wu \emph{et al}. [2021a];
the rest are from our implementations.
}
\label{main-res}
\end{table*}

\begin{table}[!t]
\centering
\resizebox{0.99\columnwidth}{!}{
\begin{tabular}{lcccc}
\toprule
 & F1$_{all}$ ($\Delta$) & 	F1$_{cross}$ ($\Delta$) & 	F1$_{intra}$ ($\Delta$) \\
\midrule
\bf POLar & 	\multicolumn{1}{l}{\bf 93.72} & 	\multicolumn{1}{l}{\bf 92.86} & 	\multicolumn{1}{l}{\bf 93.92} \\
\hline
\multicolumn{3}{l}{$\bullet$ \bf CoDiaBERT	}\\			
\quad  $\to$BERT & 	92.70 (-1.02) & 	90.75 (-2.11) & 	93.04 (-0.88) \\
\quad  w/o PSP & 	92.98 (-0.74) & 	91.28 (-1.58) & 	93.37 (-0.55) \\
\quad  PSP$\to$spk-lb & 	93.34 (-0.38) & 	92.04 (-0.82) & 	93.80 (-0.12) \\
\hdashline
\multicolumn{3}{l}{$\bullet$ \bf POLar	}\\	
\quad  w/o PGI & 	91.86 (-1.86) & 	87.28 (-5.58) & 	91.75 (-2.17) \\
\quad  w/o Pruning & 	92.25 (-1.47) & 	89.74 (-3.12) & 	92.21 (-1.17) \\
\hdashline
w/o $g_i$ (Eq. \ref{gating}) & 	93.26 (-0.46) & 	92.27 (-0.59) & 	93.50 (-0.42) \\

\bottomrule
\end{tabular}
}
\caption{Ablation results on DuConv dataset.}
\label{ablation}
\end{table}

We mainly make comparisons with the existing CSRL baselines, including CSRL \cite{XuWSZSY21}, CSAGN \cite{wu-etal-2021-csagn} and DAP \cite{wu-etal-2021-domain}.
Also we implement several representative and strong-performing models designed for regular SRL,
including UE2E \cite{LiHZZZZZ19}, LISA \cite{strubell-etal-2018-linguistically} and SynGCN \cite{marcheggiani-titov-2017-encoding}, in which we concatenate the utterances into a long sequence.
In particular, LISA and SynGCN use the external syntactic dependency trees.
Follow \newcite{XuWSZSY21}, we compute the F1 score for the detection of intra-/cross-utterance arguments (i.e., F1$_{intra}$ and F1$_{cross}$), and the overall performance (F1).

\subsection{Results and Discussions}

\paragraph{Main results.}

Table \ref{main-res} presents the main performances by different models, from which we gain several observations.
First of all, our proposed POLar system significantly outperforms all the baselines by large margins on both the in-domain and out-domain datasets, which demonstrates the efficacy of our method.
Specifically, we notice that our model achieves at least 4.07\%(=90.75-86.68) and at most 7.71\%(=45.47-37.76) F1 improvements on the cross-utterance argument detection, over the corresponding best baselines.
This significantly proves the superiority of our method on the cross-turn context modeling.
Second, by comparing the results with BERT and with CoDiaBERT, we know that our proposed CoDiaBERT PLM is of prominent helpfulness for the task.
Third, we see that with the aid of external syntactic dependency structure information, SynGCN and LISA models achieve considerable performance gains over the existing CSRL baselines (i.e., CSAGN, DAP).
However, such improvements are limited to the detection of intra-utterance arugments, contributing less to the cross-utterance arugments.
The possible reason is that, the dependency tree only works at sentence level, which fails to capture the cross-uttereance contexts.
Fortunately, our proposed latent graph can nicely compensate for this.

\begin{figure}[!t]
\centering
\includegraphics[width=1.0\columnwidth]{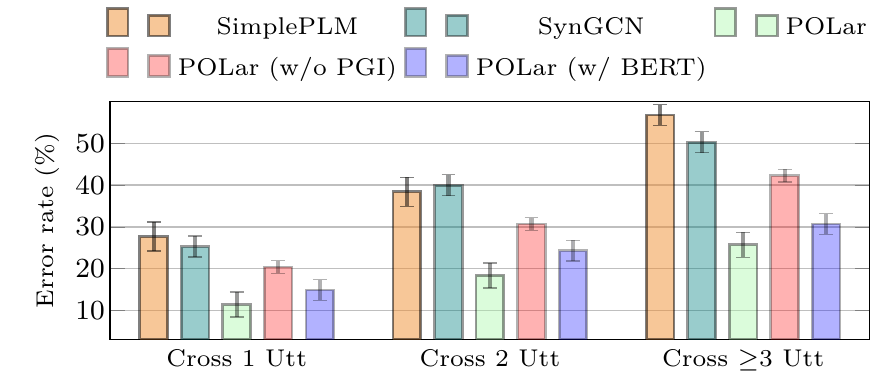}
\caption{
Error rate on cross-uttereance argument role detection.
}
\label{error-rate}
\end{figure}

\paragraph{Ablation study.}

In Table \ref{ablation} we give the model ablation results with respect to the CoDiaBERT PLM and the POLar parts, respectively.
We can observe that, by replacing the CoDiaBERT with a vanilla BERT or removing the pronoun-based speaker prediction policy (downgraded as DiaBERT), there can be considerable drops.
If we strip off the PSP, and instead use the speaker id indicator to label the speaker pronoun (i.e., spk-lb), we also witness the drops.

Further, without the PGI for the latent graph induction, i.e., directly feeding the PLM representations $\bm{h}$ in Eq. \ref{PGI-param} instead of $\bm{s}$, we can receive the most significant performance drops among all the other factors, i.e., -5.58\%F1 on the cross-utterance arguments detection.
This also reflects the importance to handle the near-neighbor influence of CSRL.
Besides, the graph pruning is quite important to the results of cross-utterance arguments.
The gating mechanism takes the positive roles to the system.

\begin{figure}[!t]
\centering
\includegraphics[width=1\columnwidth]{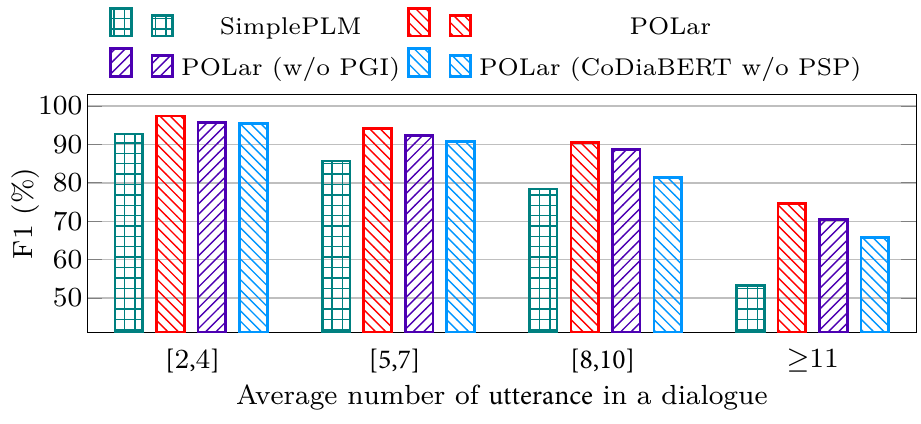}
\caption{
Influence of the utterance number in dialogue.
}
\label{utt-number}
\end{figure}

\begin{figure}[!t]
\centering
\includegraphics[width=1\columnwidth]{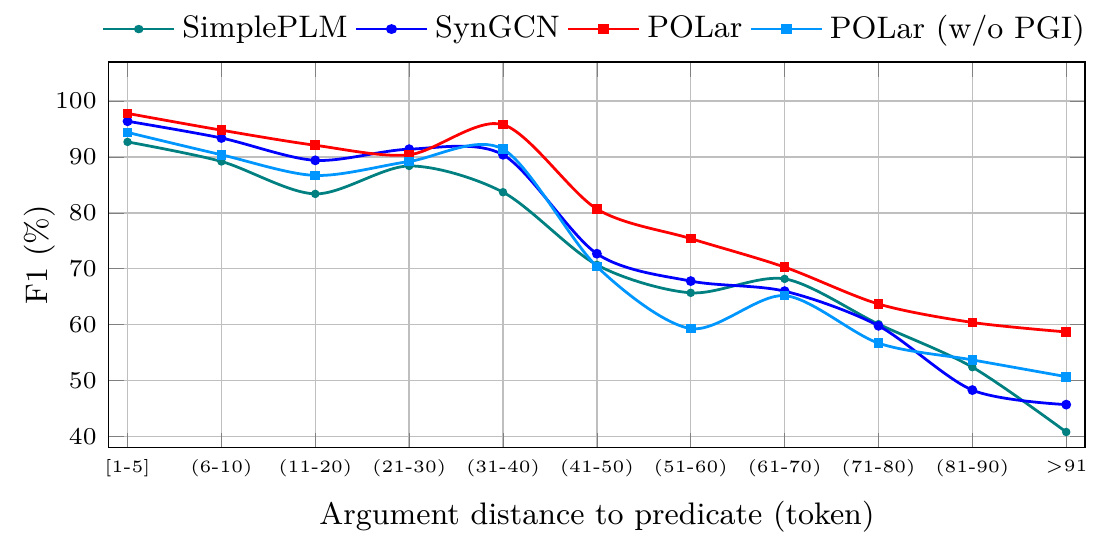}
\caption{
Influence of the argument-predicate distance.
}
\label{prd-arg-distance}
\end{figure}

\paragraph{Performances on cross-utterance argument detection.}

In Fig. \ref{error-rate} we study the error rate on the cross-utterance argument detection.
We see that with the increase of the crossed utterances, the error for the argument detection grows universally.
But in all the cases, our POLar system commits nearly half error rate, comparing to baselines.
Also we notice that, both the PGI mechanism and the CoDiaBERT is important to our system, with the former more significant than the latter.

\paragraph{Impacts of utterance numbers.}

Intuitively the more the utterance in a dialogue the severe complexity of the speaker parties, i.e., due to the speaker coreference issue.
Fig. \ref{utt-number} further plots the performances under different numbers of dialogue utterances.
It is clear that increasing the utterance number in a dialogue worsens the overall results, especially when the number $\ge$11.
In particular, the removal of PSP in CoDiaBERT shows greater impact to the removal of the PGI mechanism.
This indirectly proves that CoDiaBERT can help solve the speaker coreference issue, which gives rise to the performance gains.

\paragraph{Solving long-range dependence issue.}

Structure information has been shown effective for relieving the long-range dependence issue in SRL \cite{he-etal-2018-syntax,0001LLJ21}.
Here we explore the performances when the distances between the arguments and the predicates are different in the dialogue.
Fig. \ref{prd-arg-distance} shows that, notably, our system equipped with the latent graph performs well for those super-long argument-predicate distances, where the other baselines could fail.
Also the ablated POLar system (w/o PGI) reflects the importance of the predicate-certered Gaussian mechanism.

\paragraph{Study of the dynamic pruning for  latent graph.}

\begin{figure}[!t]
\centering
\includegraphics[width=1\columnwidth]{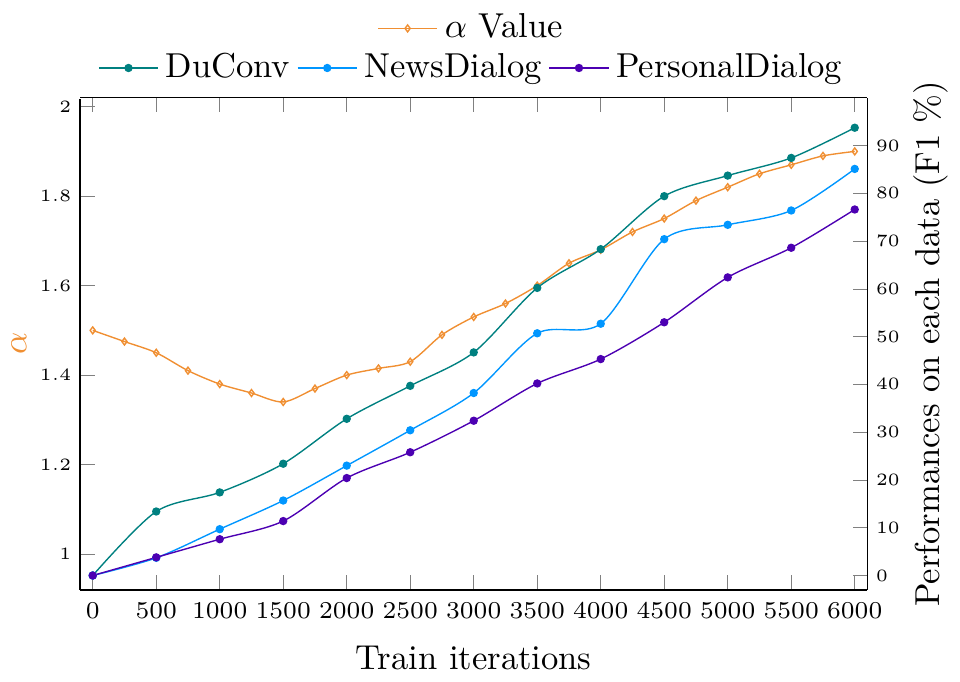}
\caption{
Trajectories of the changing pattern of $\alpha$ value, and the task performances on different data.
}
\label{alpha}
\end{figure}

Finally, we investigate the process of the dynamic pruning by studying the changing pattern of $\alpha$-Entrmax (Eq. \ref{pruning}).
Fig. \ref{alpha} plots the learning trajectories of parameter $\alpha$ as well as the variations of the correlated task performances (on three datasets).
We see that, along the training process, the $\alpha$ soon decreases to 1.35 from 1.5 at step 1,500, and then grow to 1.9, during which the latent graph becomes dense and then turns sparse gradually.
At the meantime, the CSRL performances climb to the top slowly.
This suggests that the dynamic pruning process improves the quality of POLar, which helps lead to better task demand of structure.

\section{Conclusions}

In this work we investigate the integration of a latent graph for conversational semantic role labeling.
We construct a predicate-oriented latent graph based on the two-parameter HardKuma distribution, which is induced by a predicate-centered Gaussian mechanism.
The structure is dynamically pruned and refined to best meet the task need.
Also we introduce a dialogue-level PLM for yielding better conversational text representations, e.g., supporting multiple utterance sentences, and being sensitive to the speaker roles.
Our system outperforms best-performing baselines with big margins, especially on the cross-utterance arguments.
Further analyses demonstrate the efficacy of the proposed latent graph as well as the dialogue-level PLM, respectively.
Automatically inducing task-oriented latent structure features for the structural parsing tasks is promising, which we leave as a future work.

\section*{Acknowledgments}

This work is supported by the National Natural Science Foundation of China (No.61772378, No. 62176187), 
the National Key Research and Development Program of China (No. 2017YFC1200500), 
the Research Foundation of Ministry of Education of China (No.18JZD015),
the Key Project of State Language Commission of China (No. ZDI135-112) and
the Science of Technology Project of GuangZhou (No. 20210202607).

\bibliographystyle{named}
\bibliography{main}

\end{document}